		\documentclass[preprint,12pt]{elsarticle}
		
		\usepackage{amsmath, amsfonts, amssymb, amsthm}
		\usepackage{geometry}
		\usepackage{booktabs}
		\usepackage{float}
		\usepackage{multirow}
		
		\journal{journal}
		
\begin{document}
		
		\begin{frontmatter}
		
		
		
		\title{Alternate Loss Functions for Classification and Robust Regression Can Improve the Accuracy of Artificial Neural Networks}
		
		
		\author [label1]{Mathew Mithra Noel}
		\author [label2]{Arindam Banerjee}
		\author [label3]{Yug Oswal}
		\author[label4]{Geraldine Bessie Amali D}
		\author[label5]{Venkataraman Muthiah-Nakarajan}

		\affiliation[label1]{organization={School of Electrical Engineering, Vellore Institute of Technology}, addressline={Email: mathew.m@vit.ac.in}}
		             
		\affiliation[label2]{organization={Senior Data Scientist, Ernst and Young GDS},   addressline={Email: arindam.banerjee3@gds.ey.com}}
		
		 \affiliation[label3]{organization={School of Computer Science and  Engineering, Vellore Institute of Technology}, addressline={Email: yoswal071@gmail.com}}
		
		\affiliation[label4]{organization={School of Computer Science and  Engineering, Vellore Institute of Technology}, addressline={Email: geraldine.amali@vit.ac.in}}

        \affiliation[label5]{organization={School of Electrical Engineering, Vellore Institute of Technology}, addressline={Email: mnvenkataraman@vit.ac.in}}

		\begin{abstract}
		
		All machine learning algorithms use a loss, cost, utility or reward function to encode the learning objective and oversee the learning process. This function that supervises learning is a frequently unrecognized hyperparameter that determines how incorrect outputs are penalized and can be tuned to improve performance. This paper shows that training speed and final accuracy of neural networks can significantly depend on the loss function used to train neural networks. In particular derivative values can be significantly different with different loss functions leading to significantly different performance after gradient descent based Backpropagation (BP) training. This paper explores the effect on performance of using new loss functions that are also convex but penalize errors differently compared to the popular Cross-entropy loss. Two new classification loss functions that significantly improve performance on a wide variety of benchmark tasks are proposed. A new loss function call smooth absolute error that outperforms the Squared error, Huber and Log-Cosh losses on datasets with significantly many outliers is proposed. This smooth absolute error loss function is infinitely differentiable and more closely approximates the absolute error loss compared to the Huber and Log-Cosh losses used for robust regression. 
		
		\end{abstract}

\begin{keyword}

		Loss Function \sep Robust Regression \sep Image Classification \sep Text Classification \sep Artificial Neural Network \sep Deep Learning 
		
		
		
\end{keyword}
		
\end{frontmatter}
		
		
\section{Introduction}
Artificial Neural Networks (ANNs) are a class of universal function approximators that learn a parametrized approximation to the target function through a Gradient Descent (GD) based optimization process \cite{goodfellow2016deep}. Thus learning in ANNs is reduced to the problem of learning a finite set of real parameters namely the weights and biases in the ANN model. Good parameters that result in a good approximation to the target function are computed by minimizing a loss (also known as cost) function that provides a measure of the difference between ANN outputs and targets. An important aspect of the loss function is that it distills the performance of an ANN model over a dataset into a single scalar value. The loss function is a single continuously differentiable and usually convex real-valued function of all the parameters of the ANN model. The loss function must be continuously differentiable for GD to work. Also convex loss functions are preferable since convex functions have a single global minimum. 

It is reasonable to choose a loss function inspired by statistical estimation theory to compute the most probable parameters given the dataset. Historically the Mean Square Error (MSE) and Cross-entropy from Maximum Likelihood Estimation theory (MLE) are almost universally used to train ANNs. MLE theory assigns to unknown parameters values that maximize the probability of observing the experimental data. The logarithm of the probability of observing the data is often used for convenience and results in the Cross-entropy loss. Thus MLE estimation proceed by maximizing the logarithm of the probability of the observed data as a function of the unknown parameters. For historic reasons, the negative logarithm of the probability is minimized using GD. 

The MSE is most frequently used in regression analysis where continuous real values have to be predicted and Cross-entropy is used for classification problems where the predicted outputs are discrete \cite{hastie2001elements}. Both MSE and Cross-entropy loss functions are derived from MLE theory as already described. MSE is a maximum-likelihood estimate when the model is linear and the noise is Gaussian. However it is frequently used in practice even when these assumptions cannot be justified. For classification problems both MSE and Cross-entropy losses can be used, but learning is generally faster with Cross-entropy as the gradient is larger due to the log function in Cross-entropy loss. The Cross-entropy loss also referred to as Binary Cross Entropy(BCE) loss is inspired by Information theory and measures the difference between two probability distributions \cite{brownlee2019probability}.  

\subsection{Mathematical Formalism}
Consider a dataset consisting of N training examples: $D = \{ x^i,y^i: i = 1 .. N \}$. Where $x \in R^m, y \in R^n$ in general for some natural numbers $m$ and $n$. In the following we motivate and discuss possible new loss functions for classification problems.

For binary classification problems $y \in \{0,1\}$. For the binary classification problem, the final layer consists of a single sigmoidal neuron which outputs the probability $P(y=1|x)$. This probability depends on all the weights and biases in the ANN and is hence denoted by $h_\theta(x)$. To compute the unknown parameters using maximum likelihood estimation theory the probability of the data given the parameters should be first computed.

Assuming that each training pair in the dataset is independent: \[ P(D|\theta) = \prod_{i=1}^{N} P(x^i,y^i;\theta) \].

$\prod_{i=1}^{N} P(x^i,y^i;\theta) = \prod_{i=1}^{N} P(y^i|x^i;\theta)P(x^i)$ by the definition of conditional probability.

Since $P(x^i)$ is independent of $\theta$, maximizing $P(D|\theta)$ is same as maximizing $\log(\prod_{i=1}^{N} P(y^i|x^i;\theta)) $.

Thus the log-likelihood function to be maximized is \[ \log(\prod_{i=1}^{N} P(y^i|x^i;\theta)) = \sum_{i=1}^{N} P(y^i|x^i;\theta) \]

But $P(y^i|x^i;\theta) = {h_\theta(x^i)}^{y^i}{(1-h_\theta(x^i))}^{1-y^i}$. 
 \\

Instead of maximizing the log-likelihood, the negative log-likelihood can be minimized. \\

This  negative log-likelihood after simplification is the loss \[ \mathcal{L}(\theta) = -\sum_{i=1}^{N} (y^i\log(h_\theta(x^i)) + (1-y^i)\log(1-h_\theta(x^i))) \]. \\

Scaling with any positive number does not change the minimum, so for numerical convenience we can divide by N to obtain 

\[ \mathcal{L}(\theta) = -\frac{1}{N}\sum_{i=1}^{N} (y^i\log(h_\theta(x^i)) + (1-y^i)\log(1-h_\theta(x^i)))) \]

\begin{equation} \label{eq:BCE}
l(\theta) = -\frac{1}{N}(\sum_{i=1}^{N} y^i\log(\hat{y}^i) + (1-y^i)\log(1-\hat{y}^i)) 
\end{equation}

For notational convenience $h_\theta(x)$ which is an approximation to the target $y$ is denoted by $\hat{y}$. The above binary Cross-entropy loss for a single training pair $(x,y)$ is 

\begin{equation} \label{eq:BCE_single}
l(y,\hat{y}) = -[y\log(\hat{y}) + (1-y)\log(1-\hat{y})]
\end{equation} \\

\begin{figure}[H]
\begin{center}
\includegraphics[width=7cm]{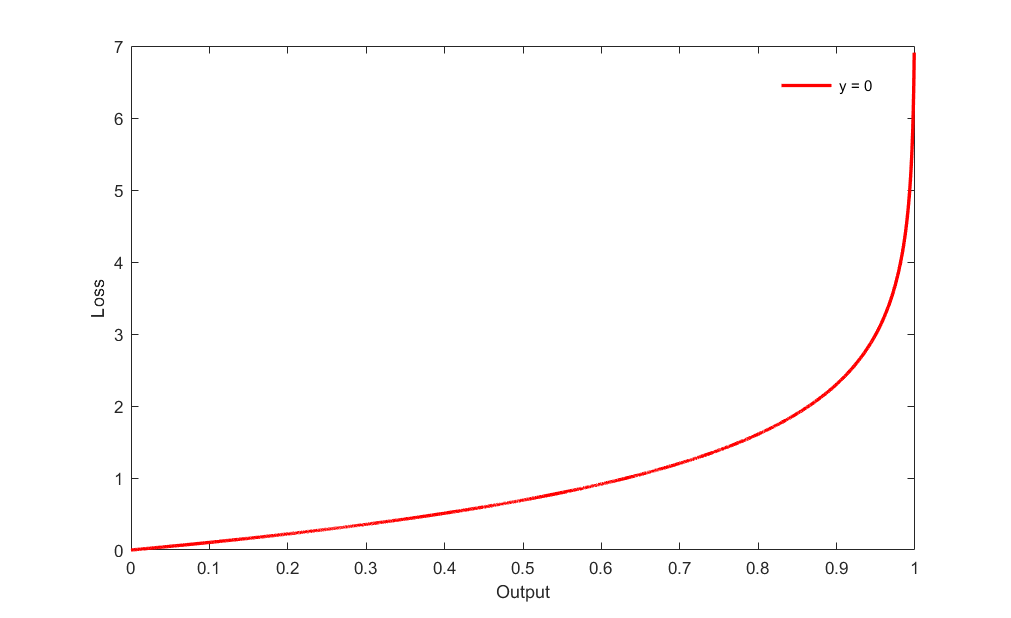}
\includegraphics[width=7cm]{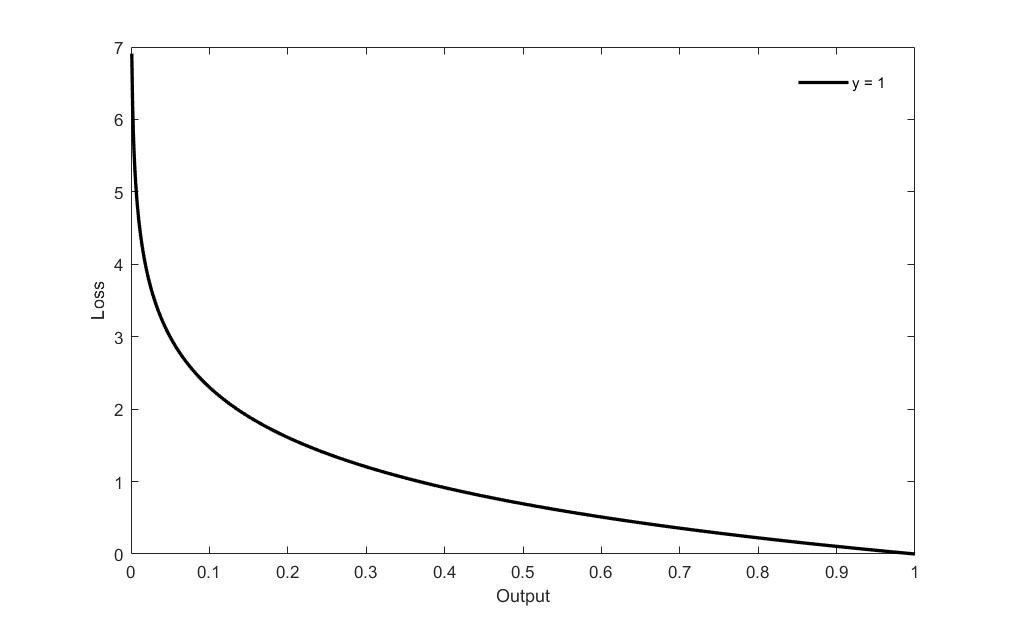}
\caption{Plot of Binary Cross-entropy loss when the target $y = 0$ (left) and $y = 1$ (right).}
\end{center}
\label{BCE}
\end{figure}

The target assumes exactly one of two possible values namely 0 or 1, so only one term in Eq.   \ref{eq:BCE_single} is nonzero. When $y=1$, $l(y,\hat{y}) = -\log(\hat{y})$ and when  $y=0$, $l(y,\hat{y}) = -\log(1-\hat{y})$. Figure \ref{BCE} shows the variation of BCE loss with binary classifier output $\hat{y}$ when the target y is 0 and 1 respectively. It is observed that the loss is exactly zero when target is equal to the output. Further the loss is differentiable, convex and tends to infinity as the output approaches the incorrect target value. Convexity is desirable to avoid introducing unnecessary local minima and differentiability is required for gradient based BP learning. In the following we propose new loss functions with the following desirable attributes:

\begin{itemize}
\label{properties}
\item $l(y,\hat{y})=0$ and $\hat{y} = y$
\item $ l(0,\hat{y}) $ and $ l(1,\hat{y})$ are convex and differentiable functions of $ \hat{y} $
\item $ l(0,0) = l(1,1) = 0 $ 
\item $ \lim_{\hat{y} \to 1} l(0,\hat{y}) = \infty $ and $ \lim_{\hat{y} \to 0} l(1,\hat{y}) = \infty $
\end{itemize}

In the above $ 0 < \hat{y} < 1 $ since $\hat{y}$ is the probability $ P(y=1|x) $. For multiclass classification problems $\hat{y}$ will be a vector with as many dimensions as classes. For C classes, the final layer will a softmax layer with C neurons and the loss function is the sum of the loss functions for each output: $ -\frac{1}{N}\sum_{j=1}^C \sum_{i=1}^{N} [y_j^i\log(\hat{y_j}^i) + (1-y_j^i)\log(1-\hat{y_j}^i)] $. In addition to using better loss functions, state-of-the-art activation functions that are known to perform better than popular activation functions like ReLu can be also used \cite{noel2021growing}, \cite{noel2023biologically} to improve performance.

During training each parameter in an ANN model is updated using $\frac{\partial \mathcal{L}}{\partial \Theta}$. Each parameter influences the loss only indirectly through the output of the network. Using the Chain rule:

\begin{equation}
	\frac{\partial \mathcal{L}}{\partial \Theta} =\frac{\partial \mathcal{L}}{\partial\hat{y}}.\frac{\partial\hat{y}}{\partial \Theta}
\end{equation}

Therefore, the derivative of the loss with respect to the output is more important than the actual loss value. To compare different loss functions we introduce the following definition: \\

\textbf{Definition}: Consider two loss functions $\mathcal{L}_1(\hat{y},y)$ and $\mathcal{L}_2(\hat{y},y)$. We define $\mathcal{L}_1$ to be stricter than $\mathcal{L}_2$ for some set of values of $ \hat{y} $ if $ \frac{\partial \mathcal{L}_1}{\partial\hat{y}}  \geq \frac{\partial \mathcal{L}_2}{\partial\hat{y}} $. \\
		
We define $\mathcal{L}_2$ to be more lenient than $\mathcal{L}_1$,  if $\mathcal{L}_1$ is stricter than $\mathcal{L}_2$. In the following we introduce  alternate loss functions that penalize network errors differently than BCE. \\

The main contributions of this work are:
\begin{itemize}

		    \item Two new loss functions that provide better performance than Cross-entropy on benchmark classification tasks are proposed
		    
		    \item A new loss function for robust regression that outperforms the Huber and Log-Cosh losses is proposed
		    
		    \item An extensive comparison of different loss functions on a variety of architectures and benchmarks is presented
		  		   
\end{itemize}

\section{New loss functions for classifciation}
Next we explore alternate loss functions that satisfy properties listed in \ref{properties}. Considering the binary classification problem and assuming that $0 < \hat{y} < 1$ and $y \in \{0,1\}$, the following new loss functions are proposed:

\begin{itemize}
\item \textbf{M Loss:} $ M(y,\hat{y}) = y\left(\frac{1}{\hat{y}}-1\right) $ \\
\item \textbf{L Loss:} $ L(y,\hat{y}) = \frac{y}{\sqrt{1-(1-\hat{y})^2}}-1 $ \\
\end{itemize}

In the above, the loss is zero if $y = 0$ and the ANN is penalized for errors only when $y = 1$. Modified versions of the above losses that penalize the ANN for errors even when $y = 0$ called full losses are given below: \\ \\

\textbf{Full loss functions:}
\begin{itemize}
\item \textbf{M Loss:} $ M(y,\hat{y}) = \frac{y}{\hat{y}} + \frac{1-y}{1-\hat{y}}-1 $ \\
\item \textbf{L Loss:} $ L(y,\hat{y}) = \frac{y}{\sqrt{1-(1-\hat{y})^2}}+\frac{1-y}{\sqrt{1-\hat{y}^2}}-1 $ \\
\end{itemize}

\begin{figure}[H]
\begin{center}
\includegraphics[scale=0.5]{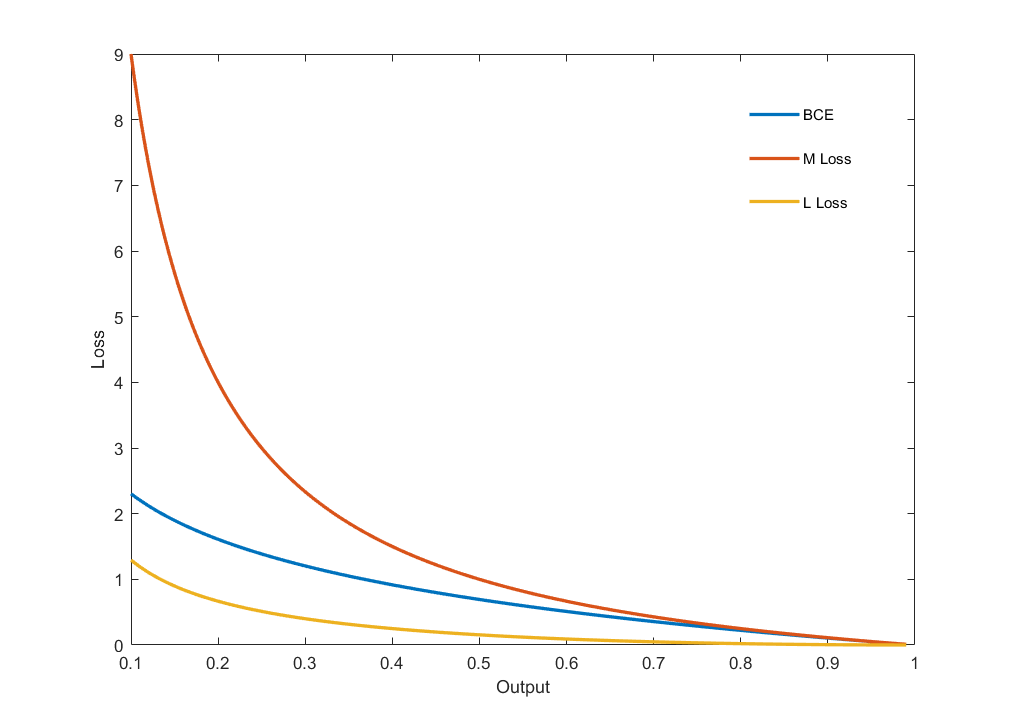}
\end{center}
\caption{Plot of different loss functions when the target y = 1.}
\label{fig:functions}
\end{figure}

Figure \ref{fig:functions} shows the difference between Cross-entropy and other loss functions proposed in this paper. It is observed from Fig. \ref{fig:functions} that the M loss is stricter than BCE and that the L loss is more lenient than BCE for most output values. To apply the above loss functions to multiclass classification problems, the losses for each output are added together. In particular if $l(y_j^i,{\hat{y}_j}^i)$ is the loss for the $j^{th}$ output and $i^{th}$ training pair, the overall loss is $ \mathcal{L} = \frac{1}{N} \sum_{j=1}^C \sum_{i=1}^{N} l(y_j^i,{\hat{y}_j}^i) $ where N is the size of the mini-batch for mini-batch gradient descent.

\section{New loss functions for robust regression}

For Regression problems the Mean Square Error (MSE) is most frequently used. However the MSE is very sensitive to outliers, and a differentiable approximation to the Mean Absolute Error (MAE) is desired. In particular if the training data for regression comes from a disctribution with a "heavy tail", the presence of many outliers will lead to large prediction errors if MSE is used during training. The MAE weights data equally and is robust to the presence of outliers, however its use is limited since it is not differentiable at the origin. Another problem with the MAE loss is that it has a large derivative ($ \pm 1 $) close to the origin leading to oscillations about the minimum during gradient descent. ANN training loss functions have to be differentiable, since the Backpopagation algorithm requires the loss function to be differentiable. Thus differentiable and computationally cheap alternatives to the MSE loss like the Huber loss \cite{10.1214/aoms/1177703732} and logcosh loss \cite{wang2020comprehensive} have been explored in the past. This paper poposes the following computationally cheap  infinitely differentiable smooth approximation to the MAE loss called Smooth MAE (SMAE):
 \[ SMAE(e) = e \tanh(\frac{e}{2}) \]
 
 The SMAE loss proposed in this paper has the following nice properties:
 
 For small error values ($|e| << 1$), SMAE approximates the MSE, since $SMAE(e) = e \tanh(\frac{e}{2}) \approx \frac{e^2}{2}$.
 
 For large error values ($e \rightarrow \infty$), SMAE approximates the MAE, since $SMAE(e) = e \tanh(\frac{e}{2}) \approx |e|$.

\begin{figure}[H]
	\begin{center}
		\includegraphics[scale=0.5]{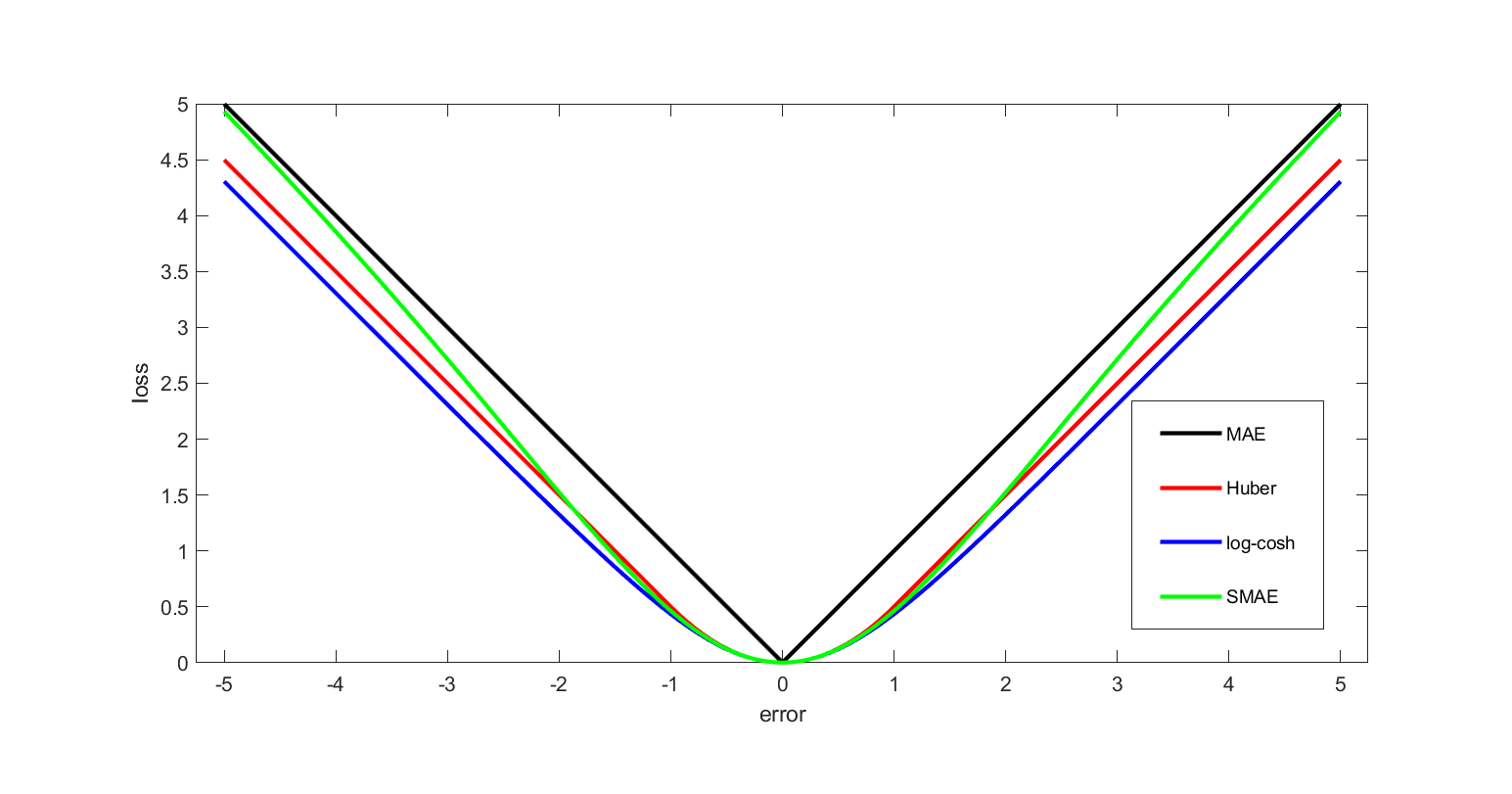}
	\end{center}
	\caption{Plot of different loss functions used for robust regression.}
	\label{reg_losses}
\end{figure}

Figure \ref*{reg_losses} compares the SMAE loss proposed in this paper with different loss functions used for regression. It is clear from Figure \ref*{reg_losses} that SMAE approximates the ideal MAE loss more accurately than other loss functions used for robust regression. Also it is clear from Fig. \ref*{reg_losses} that SMAE loss converges to the MAE loss for large error values unlike the Huber and Log-Cosh losses that have a permanent offset with respect to the ideal MAE loss. Further the Huber loss has a hyperparameter that has to be tuned, so the Huber loss is not directly comparable to other loss functions without hyperparameters. 

\begin{figure}[H]
	\begin{center}
		\includegraphics[scale=0.5]{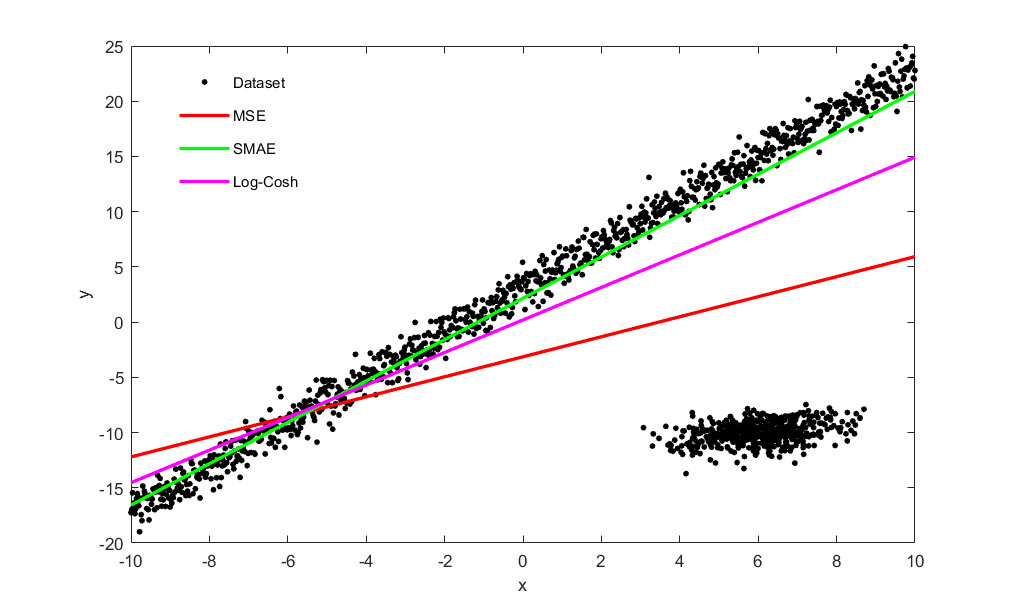}
		
	\end{center}
	\caption{Effect of using different loss functions for linear regression when the dataset has significant number of outliers. The training dataset is shown as black dots and consists of 1500 instances with 500 outliers.}
	\label{reg_plots}
\end{figure}

Figure \ref{reg_plots} shows that the SMAE loss proposed in this paper is less affected by outliers than other loss functions. The robustness of the SMAE loss can be attributed to its close approximation to the ideal MAE loss compared to other loss functions. Table \ref{estimation} shows the percentage parameter estimation error when different loss functions are used to estimate the parameters of a linear model. The training data consisted of 500 outliers and 1000 points(Fig. \ref{reg_plots}) generated by adding zero mean unit variance Gaussian noise to a linear model $y = 2*x+3$. The 500 outliers were generated from a 2D Gaussian distribution with mean vector $\mathbf{\mu}$ and covariance matrix $\mathbf{\Sigma}$ shown below:

$$ \mathbf{\mu} = 
\begin{bmatrix}
	  6\\
	-10\\
\end{bmatrix} $$

$$ \mathbf{\Sigma}=
\begin{bmatrix}
	 1 & 0.3\\
	0.3 & 1\\
\end{bmatrix} $$

\begin{table}[H]
	\centering
	\caption{Parameter estimation accuracy with different regression loss functions on a dataset with outliers.}
	\begin{tabular}{|c|c|c|}  \hline
		\centering
		\textbf{Loss} & \multicolumn{2}{c|}{\textbf{Parameter Estimation Error (\%)}}  \\ \cline{2-3}
		\textbf{Function} & Slope & Y-intercept \\  \hline
		SMAE & 8.73 & 22.6 \\
		MSE & 68.2 & 205.82\\
		Log-Cosh & 22.76 &	82.3\\ \hline
	\end{tabular}
	\label{estimation}
\end{table}

\section{Comparison of performance on benchmark vision datasets}
\label{section:results}

In the following the two different versions of the M loss and L loss functions are compared to the standard cross-entropy loss on a variety of benchmark architectures and problems. The different loss functions are only used during training to compute the parameters of the model. After training the performance of different models was compared using the accuracy on a separate test dataset as the metric.

\subsection{Comparison on the CIFAR-10 benchmark}
The different loss functions were used to train the VGG-19 model \cite{simonyan2015deep} with the CIFAR-10 dataset \cite{krizhevsky2009learning}. The CIFAR-10 dataset consists of 60000, 32x32 color images organized into 10 classes, with 6000 images per class. There are 50000 training images and 10000 test images.The Adam optimizer with learning rate of 1e-5 was used. The model was trained on a Google Colab environment with 50 epochs and 100 steps per epoch. Five loss functions - i) Cross Entropy loss, ii) L loss, iii) M-loss, iv) Full L loss, and v) Full M-loss were considered. The training and test accuracy values for these five loss functions are shown in Table \ref{table1}.

 \begin{table}[H]
	\centering
	\caption{Performance of VGG-19 trained using different loss functions on the CIFAR-10 benchmark}
	\begin{tabular}{|c|c|c|} \hline
		\textbf{Loss Function} & \textbf{Training Accuracy} & \textbf{Test Accuracy} \\ \hline
		Cross Entropy loss &	0.5806 &	0.5668  \\
		L loss &	0.5791 &	0.5684  \\
		M loss &	0.5745 &	0.5666  \\
		Full L loss &	0.5827 &	0.571  \\
		Full M loss	&0.5762&	0.5668  \\ \hline
	\end{tabular}
	\label{table1}
\end{table}

 \begin{figure}[H]
 	\begin{center}
 	\includegraphics[width=0.8\linewidth]{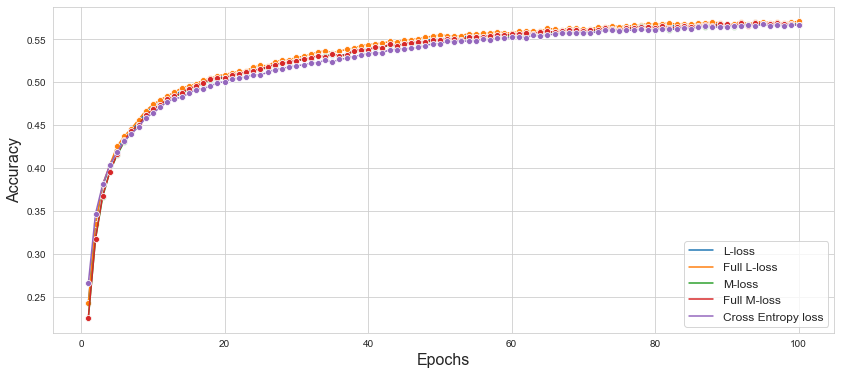}
 	\end{center}
 	\caption{Variation of accuracy with epochs on the CIFAR-10 benchmark.}
 	\label{acc1}
 \end{figure}

 \begin{figure}[H]
 	\begin{center}
 		\includegraphics[width=0.8\linewidth]{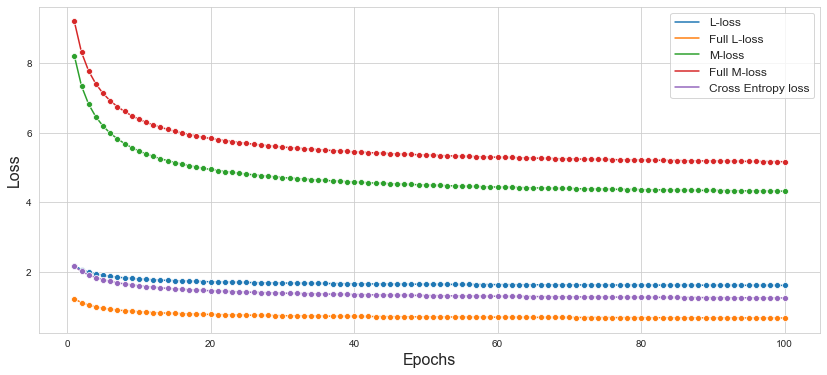}
 	\end{center}
 	\caption{Variation of loss with epochs for the CIFAR-10 benchmark.}
 	\label{loss1}
 \end{figure}

 Figures \ref{acc1} and \ref{loss1} show the progression of training with different loss functions. It is observed that the full L loss outperforms all other loss functions.
 
 \subsection{Comparison on Imagenette Dataset}
 
 The different loss functions were used to train the VGG-19 model with the Imagenette dataset \cite{howard2020imagenette}. Imagenette is a subset of the Imagenet dataset and consists of 10 classes. The 160px-v2 version of imagenette has 9,469 training images and 3,925 test images. For transfer learning, the VGG-19 pre-trained model from Tensorflow having pre-trained “imagenet” weights was used. The top fully connected layers were not included and were explicitly defined to be trained from the given dataset. The Adam optimizer with a learning rate of 1e-5 was used for training the model. The model was trained on a Colab environment with 50 epochs and 100 steps per epoch. The performance of the 5 loss functions on the Imagenette benchmark is presented in Table \ref{Imagenette}.

 \begin{table}[H]
 	\centering
 	\caption{Performance of VGG-19 trained using different loss functions on the Imagenette benchmark} 
 	\begin{tabular}{|c|c|c|}   \hline
 		\textbf{Loss Function} & \textbf{Training Accuracy} & \textbf{Test Accuracy}  \\ \hline
 		Cross Entropy loss &	0.73	&0.74\\
 		L loss	&0.75 &	0.77\\
 		M loss	&0.74 &	0.78\\
 		Full L loss &	0.72 &	0.78\\
 		Full M loss	& 0.74 &	0.76		 \\ \hline
 	\end{tabular}
 	\label{Imagenette}
 \end{table}
 
\begin{figure}[H]
 	\begin{center}
 		\includegraphics[width=0.8\linewidth]{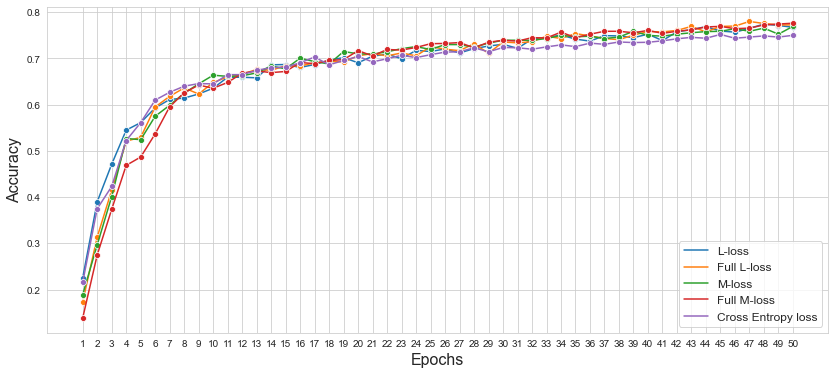}
 	\end{center}
 	\caption{Variation of accuracy with epochs for the Imagenette benchmark.}
 	\label{Imagenette1}
 \end{figure}

\begin{figure}[H]
 	\begin{center}
 		\includegraphics[width=0.8\linewidth]{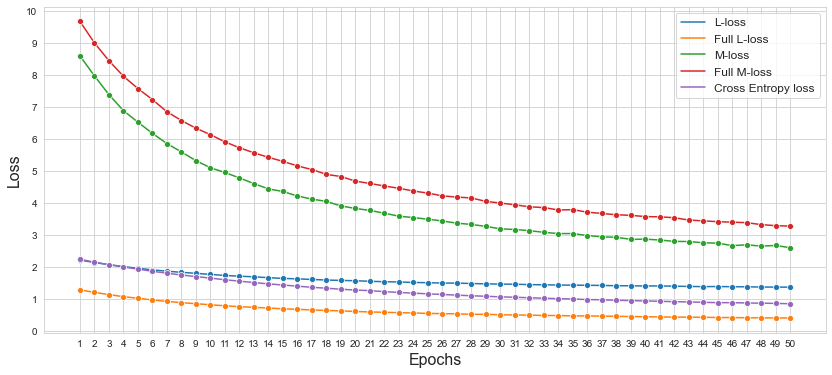}
 	\end{center}
 	\caption{Variation of loss with epochs for the Imagenette benchmark.}
 	\label{Imagenette2}
 \end{figure}

 Table \ref{Imagenette} shows that the test accuracy with M loss and L losses (both versions) are significantly higher than Cross Entropy loss. Figures \ref{Imagenette1} and \ref{Imagenette2} show the progression of training with different loss functions. It is observed that the full L loss proposed in this paper outperforms all other loss functions on the Imagenette benchmark.

\section{Comparison on text datasets}

In order to test the generality of the proposed loss functions beyond computer vision tasks and the VGG-19 model, text based benchmarks and advanced architectures were considered. 

\subsection{Comparison on Consumer Financial Protection Bureau (CFPB) dataset}
First a LSTM based model was used to classify text from the Consumer Financial Protection Bureau (CFPB) dataset \cite{CFPB}. Each week the CFPB sends thousands of consumers’ complaints about financial products and services to companies for response. Those complaints are published after the company responds or after 15 days, whichever comes first. In the CFPB dataset the target variable has 10 possible classes. The model contains an embedding layer followed by a spatial dropout, an LSTM layer of 100 units, and a dense layer to output 10 possible classes. 

\begin{table}[H]
\centering
\caption{Performance of a LSTM model trained using different loss functions on the CFPB benchmark}
\begin{tabular}{|c|c|} \hline
\textbf{Loss Function} & \textbf{Test Accuracy}\\ \hline 
BCE loss & 0.822 \\
L loss & 0.843\\
M loss & 0.827\\
Full L loss & 0.839\\
Full M loss & 0.825\\ \hline
\end{tabular}
\label{LSTM}
\end{table}

The test accuracy of the LSTM model when trained using different loss functions is shown in Table \ref{LSTM}.These experimental results show that both variants of the proposed L loss and M loss outperform the cross-entropy loss in most of the cases.

\subsection{Comparison on the IMDB Movie Review benchmark}
All 5 loss functions were used to train a BERT sentiment analysis model to classify movie reviews as positive or negative, based on the text of the review. The Large Movie Review Dataset \cite{maas-EtAl:2011:ACL-HLT2011} that contains the text of 50,000 movie reviews from the Internet Movie Database (IMDB) was used to train the BERT model. The Smaller BERT model from TensorFlow Hub having 4 hidden layers (Transformer blocks), a hidden size of 512, and 8 attention heads. Adam optimizer was used with this BERT model was considered. The test accuracy values for these five loss functions are shown in Table \ref{BERT}. Table \ref{BERT} indicates that the L loss and full L loss outperform all other losses by a small margin.

\begin{table}[H]
	\centering
	\caption{Performance of BERT model trained using different loss functions on the IMDB benchmark.}
	\begin{tabular}{|c|c|}  \hline
		\textbf{Loss Function} & \textbf{Test Accuracy}\\ \hline
		BCE loss &	0.8551 \\
		L loss &	0.8557  \\
		M loss &	0.8539  \\ 
		Full L loss &	0.8577 \\
		Full M loss	& 0.8555  \\ \hline 
	\end{tabular}
	\label{BERT}
\end{table}

\section{Conclusion}
This paper explored the possible advantages of using alternatives to popular loss functions for classification and robust regression. Four new loss functions for classification were proposed and shown to outperform the popular BCE loss on a variety of architectures and benchmarks. The losses for classification proposed in this paper were shown to outperform the popular BCE loss on all benchmarks and architectures considered. On certain tasks the proposed loss functions were observed to significantly increase the classification accuracy. The M loss and full M loss were stricter than the BCE loss in penalizing errors. The L and full L-losses were lenient compared to the BCE loss. The superior performance of the L and M losses indicate that the BCE loss inspired by MLE theory is not optimal and better loss functions can be designed.

Many real-world datasets are contaminated with significant numbers of large outliers and these outliers can confuse regression algorithms leading to large prediction errors. This paper proposes a very close infinitely differentiable smooth computationally cheap approximation to the Mean Absolute Error (MAE) loss referred to as the SMAE loss. The SMAE loss does not require the choice of an additional hyperparameter like the Huber loss used in robust regression and is significantly more robust to outliers than the Log-Cosh loss.

\bibliographystyle{elsarticle-num}
\bibliography{bibliography.bib}

		
\end{document}